\documentclass[]{article} 

\usepackage{amssymb}
\usepackage{amsmath}
\usepackage{graphicx}
\usepackage{subfigure}
\usepackage{makeidx}
\usepackage{multicol}
\usepackage[export]{adjustbox}
\usepackage[justification=centering]{caption}
\usepackage{chngcntr}
\counterwithout{figure}{section}
\usepackage{float}

\usepackage{comment}
\usepackage{amsmath}
\usepackage{algorithm}
\usepackage{algorithmic}
\usepackage{caption}

\usepackage{adjustbox}
\usepackage{multirow}
\usepackage{stfloats}
\usepackage{booktabs}
\usepackage{siunitx}
\usepackage{arydshln}
\usepackage{url}
\usepackage{ragged2e}
\usepackage{appendix}
\usepackage{xcolor} % 色を使うために必要
\def\B#1{\textcolor{black}{#1}}

\def\R#1{#1} % カメラレディ用の黒に変更

\usepackage{fancyhdr}
\let\origtitle\title 
\renewcommand{\title}[1]{\lfoot{\textit{#1}}\origtitle{\textbf{#1}}}
\renewcommand{\sectionmark}[1]{\markboth {}{}}

\date{}

\title{Self-supervised Learning Method Using Transformer for Multi-dimensional Sensor Data Processing}
\begin{document}
\maketitle
\thispagestyle{fancy}
\centering

\B{\author{
Haruki Kai \footnote{kai.haruki822@mail.kyutech.jp},  
Tsuyoshi Okita \footnote{tsuyoshi.okita@gmail.com}}}\\
\thanks{$^1$$^2$Kyushu Institute of Technology}
%, $^3$$^5$$^7$Hokkaido University, $^4$Hiroshima International University, $^6$Kagawa University, $^8$JIKEI University}

%%%%%%%% abstract from here %%%%%%%%%%
\abstract{
%We developed a deep learning algorithm for human activity recognition using sensor signals as input. In this study, we constructed a pre-trained language model based on the Transformer architecture commonly employed in natural language processing. Leveraging this pre-trained language model, we aimed to address the downstream task of human activity recognition. While this can be achieved with a vanilla Transformer, we propose a novel n-dimensional numerical processing Transformer that incorporates three distinctive features: embedding n-dimensional numerical data using a linear layer, binning preprocessing, and linear transformation in the output layer. The effectiveness of the proposed model was evaluated on five different datasets. Compared to the vanilla Transformer, our model achieved a 10\%–15\% improvement in accuracy.
We developed a deep learning algorithm for human activity recognition using sensor signals as input. In this study, we built a pre-trained language model based on the Transformer architecture, which is widely used in natural language processing. By leveraging this pre-trained model, we aimed to improve performance on the downstream task of human activity recognition. While this task can be addressed using a vanilla Transformer, we propose an enhanced n-dimensional numerical processing Transformer that incorporates three key features: embedding n-dimensional numerical data through a linear layer, binning-based preprocessing, and a linear transformation in the output layer. We evaluated the effectiveness of our proposed model across five different datasets. Compared to the vanilla Transformer, our model demonstrated a 10\%–15\% improvement in accuracy.
}

\section{Introduction}
\label{section:Introduction}

In recent years, Human Activity Recognition (HAR), which involves inferring human activities from sensor data, has gained increasing importance across various domains, including healthcare, sports, and smart home applications \cite{lara2012survey, chen2021deep, 10651718, 10652200, AkihisaTsukamoto202426, 10651886}. Traditional HAR methods often rely on camera-based approaches, which have been extensively studied. However, these methods present challenges such as environmental constraints and privacy concerns for individuals being monitored \cite{jung2020review}. In contrast, approaches based on motion sensor data—such as accelerometers and gyroscopes embedded in smartphones and wearable devices—offer significant advantages, including reduced dependency on environmental conditions and improved privacy protection.

With advancements in IoT technologies and the widespread adoption of smart devices, it has become increasingly feasible to collect and store large-scale multi-dimensional time-series data from sensors like accelerometers and gyroscopes \cite{MONDAL2021343}. HAR methods leveraging such sensor data not only offer enhanced privacy but also hold great potential for applications such as daily activity monitoring and healthcare services.

This study aims to achieve high-accuracy activity classification using multi-dimensional time-series data from sensors such as accelerometers and gyroscopes. Specifically, we propose an algorithm with the following key features:

\begin{itemize}
    \item \textbf{Embedding for Multi-Dimensional Numerical Data}:  A simple linear layer is introduced to embed multi-dimensional numerical data into vector representations.
    \item \textbf{Binning Process}: Time-series data is discretized to mitigate noise and enhance feature extraction.
    \item \textbf{Parallel Linear Layers in the Output}: Parallel linear layers are designed to accommodate varying channel characteristics, thereby improving prediction accuracy.
\end{itemize}
\begin{comment}
    Although the application of Transformer\cite{transformer} to HAR tasks has been gradually increasing, many studies focus on designing custom Transformer models specifically tailored for sensor data \cite{transformerinfluenza, informer}. On the other hand, in the field of natural language processing (NLP), various Transformer-based models, such as BERT \cite{bert}, the GPT series \cite{gpt-2}, and DistilBERT \cite{distilbert}, have been proposed and refined to achieve a balance between high accuracy and computational efficiency.

Based on the idea that adapting existing NLP models for sensor data analysis could incorporate state-of-the-art advancements in language modeling into HAR tasks, we attempt to build a HAR model based on NLP models. Applying these efficient Transformer architectures directly to sensor data remains relatively unexplored, and their effectiveness requires validation.

The contributions of this study are as follows:
\end{comment}

Although the application of Transformer models\cite{transformer} to HAR tasks has been steadily increasing, many existing studies focus on designing custom Transformer architectures specifically tailored for sensor data \cite{transformerinfluenza, informer}. In contrast, the field of natural language processing (NLP) has seen the development and refinement of various Transformer-based models, such as BERT \cite{bert}, the GPT series \cite{gpt-2}, and DistilBERT \cite{distilbert}, which are designed to achieve a balance between high accuracy and computational efficiency.

Inspired by the idea that adapting existing NLP models for sensor data analysis could integrate state-of-the-art advancements in language modeling into HAR tasks, we explore the feasibility of building a HAR model based on NLP-oriented Transformers. However, the direct application of these efficient Transformer architectures to sensor data remains relatively unexplored, and their effectiveness requires empirical validation.

The contributions of this study are as follows:
\begin{comment}
    \begin{itemize}
    \item \textbf{Efficient Model Design Leveraging NLP Models}: Using transformer-based models, such as DistilBERT and GPT-2, which have demonstrated success in NLP, we provide design guidelines for direct application to sensor data analysis.
    \item \textbf{Proposed Embedding, Binning, and Parallel Linear Layer Structure}: As a novel contribution, we introduce a lightweight yet flexible model that combines extensive use of linear layers for embedding, binning mechanisms for noise reduction, and parallel linear layers in the output.
    \item \textbf{Experimental Validation}: We validate the effectiveness of the proposed model by comparing it with conventional models, including ResNet, Random Forest, and Vanilla Transformer\cite{transformer}.
\end{itemize}
\end{comment}

\begin{itemize}
\item \textbf{Efficient Model Design Leveraging NLP Models}: We explore the direct application of Transformer-based models, such as DistilBERT and GPT-2, which have demonstrated success in NLP, to sensor data analysis. We provide design guidelines to effectively adapt these models for HAR tasks.
\item \textbf{Proposed Embedding, Binning, and Parallel Linear Layer Structure}: As a novel contribution, we introduce a lightweight yet flexible model that integrates linear layers for embedding, binning mechanisms for noise reduction, and parallel linear layers in the output to enhance predictive performance.
\item \textbf{Experimental Validation}: The effectiveness of the proposed model is empirically validated through comparative experiments against conventional models, including ResNet, Random Forest, and Vanilla Transformer\cite{transformer}.
\end{itemize}

\section{Related literature}
\label{section: Related literature}

In recent years, Transformer architectures have been increasingly applied beyond the field of natural language processing (NLP) to non-linguistic data analysis, including sensor data and time-series data. This trend can be attributed to the unique characteristics of Transformers, particularly their self-attention mechanism, which captures long-term dependencies while enabling parallel computation.

\subsection{\R{Soft sensing transformer}}

\R{Recent advancements in sensor-based human activity recognition (HAR) have seen the adaptation of Transformer architectures for handling multi-dimensional sensor data. For example, the Soft Sensing Transformer \cite{zhang2021soft} replaces conventional embedding layers with linear transformations to directly process continuous sensor inputs. This approach efficiently handles multi-dimensional data without requiring explicit tokenization or clustering. Our work builds on this idea by further enhancing self-supervised learning for HAR through additional design components.}

\subsection{\R{Time-Series Transformer Variants}}

\R{There are several categories of recent works for time-series
transformer.  The first category is the modification on attention
mechanism.  LogTrans \cite{Li2019} proposes convolutional
self-attention by employing causal convolutions to generate queries
and keys in the self-attention layer. LogTrans introduces sparse bias,
a Logsparse mask, in self-attention model.  Informer 
\cite{Zhou2021}
%[Zhou et al.,2021] 
selects dominant queries based on queries and key
similarities. Together with this, Informer designs a generative style
decoder to produce long-term forecasting directly.  AST 
\cite{Wu2020a}
%[Wu et al., 2020a] 
uses a generative adversarial encoderdecoder framework to train
a sparse Transformer model.  Pyraformer 
\cite{Liu2022a}
%[Liu et al., 2022a] 
designs a
hierarchical pyramidal attention module with a binary tree following
the path, to capture temporal dependencies of different ranges.
FEDformer 
\cite{Zhou2022}
%[Zhou et al., 2022] 
applies attention operation in the
frequency domain with Fourier transform and wavelet transform.
Quatformer 
\cite{Chen2022}
%[Chen et al., 2022] proposes learning-to-rotate attention
(LRA) based on quaternions that introduce learnable period and phase
information to depict intricate periodical patterns.}

\R{The second categories introduces the way to normalize time series
data.  Non-stationary Transformer 
\cite{Liu2022b}
%[Liu et al., 2022b] 
introduces the
modification mechanism in the normalization.}

\R{The third categories utilizes the bias for token input.
Autoformer 
\cite{Wu2021}
%[Wu et al.,2021] 
adopts a segmentation-based representation
mechanism.  The auto-correlation of this block measures the time-delay
similarity between inputs signal.
PatchTST 
\cite{Nie2023}
%[Nie et al., 2023] 
uses channels which
contain a single univariate time series that shares the same embedding within
all the series. And it uses subseries-level patch design which segmentation of time series into subseries-level patches that are
served as input tokens to Transformer.
Crossformer 
\cite{ZhangYan2023}
%[Zhang and Yan, 2023] 
uses the input which is embedded into a 2D vector array through the dimension-segment-wise embedding to preserve time and dimension information.}

\subsection{\R{Vision Transformer Variants}}

\R{
%SignalTransformer-MLM [anonymous\footnote{Self reference.}] 
SENvT-u4 \cite{Okita23} is a transformer model designed to handle signals, with pre-training consisting of MLM (Masked Language Modeling) and a signal-specific pretext task. 
%SignalTransformer-contrastive [anonymous\footnote{Self reference.}] 
SENvT-contrastive \cite{Okita23} is also a transformer designed for signals, but its pre-training is based on contrastive learning. Both of these models are modified versions of the original Vision Transformer (ViT). In the Vision Transformer, a grid of patches is used, with these patches treated as embeddings. Moreover, the relative relationships between the embeddings of a given patch and its adjacent patches are recognized as distances.}

\R{The signal transformers modify the basic structure of the Vision Transformer by modifying it from two-dimensional to one-dimensional. As a result, it treats signals in the time series direction as one-dimensional patches, embedding them in the same manner. However, unlike Vision Transformers, the time series signals are not treated as individual data points but as embeddings based on one-dimensional patches. These patches are treated as embeddings during input. Therefore, they do not take the form of the linear layer proposed in this paper.}

\subsection{Attempts to Leverage LLMs for Sensor Data Analysis}
More recently, attempts have been made to apply large language models (LLMs), such as GPT-4, to sensor data-based activity recognition tasks \cite{llmzeroshot}. For instance, Sijie Ji et al. \cite{llmzeroshot} designed prompts, such as "role-play scenarios" and "think step-by-step" to achieve accuracy surpassing that of traditional baseline models. This study suggests the potential of repurposing the extensive knowledge encoded in LLMs for sensor data analysis.

However, such approaches often rely on natural language tokenizers to encode sensor data, which may limit the ability to fully exploit the continuous nature of numerical data. Natural language tokenizers are inherently designed to segment strings or word sequences into discrete tokens, making them less suited for effectively representing multi-dimensional sensor data. Addressing this challenge requires the development of embedding layers tailored to numerical data or structural modifications to LLMs.

Building upon these advancements in applying Transformers to non-linguistic data analysis, this study proposes a model that combines linear layers for efficient embedding of sensor data with architectures based on natural language models. Our aim is to achieve high accuracy in activity recognition tasks by leveraging the strengths of both linear embeddings and Transformer-based architectures.

\section{Method}
\label{section:Method}
\begin{figure}[h]
    \centering
    \includegraphics[width=0.9\textwidth]{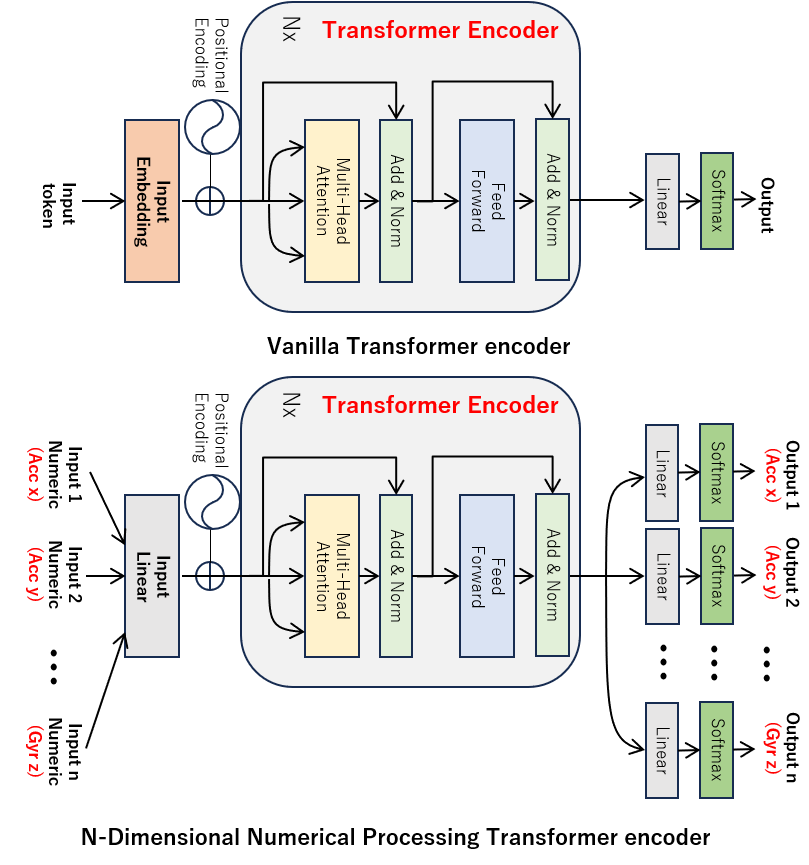}
    \caption{Conventional and proposed architectures}
    \label{fig:arch_encoder_pretrain}
\end{figure}

We propose a method for processing n-dimensional numerical data based on the Transformer encoder commonly used in natural language processing. The architecture is illustrated in Figure \ref{fig:arch_encoder_pretrain}. Specifically, the proposed method consists of the following components:\\

\textbf{Embedding with Linear Layer}:A method that takes n-dimensional numerical data as input and obtains an embedded representation through a linear layer.
In our approach, multi-dimensional sensor data is processed in a manner similar to natural language processing (NLP) models, where sequential token embeddings are fed into a Transformer.
\begin{comment}
    Specifically, given an input of shape 
\(
L \times n
\), where 
\(L\) represents the sequence length (e.g., 300 for accelerometer data) and 
\(n\) represents the number of sensor dimensions (e.g., 3 for a triaxial accelerometer), 
\end{comment}
\B{Specifically, given an input of shape \( L \times n \), where:
\begin{itemize}
    \item \( L \) represents the sequence length (e.g., 300 for accelerometer data),
    \item \( n \) represents the number of sensor dimensions (e.g., 3 for a triaxial accelerometer).
\end{itemize}}

\begin{comment}
    we apply a single linear embedding layer of size 
\(
n \times d
\) to project each 
\(n\)-dimensional sensor reading into a 
\(d\)-dimensional representation. 
\end{comment}
\B{We apply a single linear embedding layer of size \( n \times d \), where:
\begin{itemize}
    \item \( n \) represents the number of sensor dimensions,
    \item \( d \) represents the embedding dimension.
\end{itemize}}
This ensures that each sensor reading is converted into a unified feature space while preserving temporal order.
Although this paper focuses primarily on 3-dimensional sensor data (e.g., accelerometer data with X, Y, and Z axes), the proposed method is not restricted to this configuration. It can be naturally extended to higher-dimensional sensor data (e.g., 6-axis IMU with accelerometer and gyroscope, or 9-axis motion sensor data), where the linear embedding layer would be adjusted to match the input dimensionality. By embedding each time step independently using a single linear layer, the model can learn meaningful representations of sensor readings while maintaining their temporal structure, similar to how token embeddings are processed in NLP tasks.\\

\textbf{Parallel linear layers in the output layer}:A design for the output layer with parallel linear layers corresponding to each dimension of the n-dimensional data, enabling the learning of data features.
\R{A single output layer for all dimensions could, in theory, capture inter-axis dependencies; however, this would require additional preprocessing steps such as clustering or tokenization to consolidate multi-dimensional sensor readings into a unified representation. This transformation could lead to the loss of valuable information specific to individual sensor axes. Instead, our approach maintains the original structure of the sensor data by employing separate output layers for each dimension. This allows the model to independently learn the characteristics of each sensor axis while still leveraging the Transformer’s attention mechanism to capture inter-axis dependencies implicitly.}\\

\textbf{Binning process}:A discretization method for input numerical data to facilitate self-supervised learning in the output layer.\\

\textbf{Pre-Training algorithm}:A proposed Pre-Training algorithm designed for the proposed architecture.

\R{While Soft Sensing Transformer \cite{zhang2021soft} also applies linear transformations to sensor inputs, our approach further introduces a binning process for efficient self-supervised learning and parallel linear layers in the output layer to handle multi-dimensional data without requiring additional clustering or tokenization.
Through this design, we propose an effective architecture for HAR tasks that differs from similar studies in its structural approach.}

Notably, this method is not limited to Transformer encoders and can be similarly applied to Transformer decoders.

This chapter provides a detailed explanation of each of these components.

\subsection{Embedding with Linear Layer}
\label{section:Embedding with linear layer}
In traditional natural language models, an embedding layer is used to convert input data into embedding vectors. In contrast, our method replaces the embedding layer with a linear layer to transform the input data $\mathbf{x} \in \mathbb{R}^n$ into an embedding vector $\mathbf{h} \in \mathbb{R}^d$. The proposed embedding using a linear layer is defined as follows:
\[
\mathbf{h} = \mathbf{W} \mathbf{x} + \mathbf{b}
\]
\begin{itemize}
    \item $\mathbf{x} \in \mathbb{R}^n$ is the input n-dimensional numerical data,
    \item $\mathbf{W} \in \mathbb{R}^{d \times n}$ is a learnable weight matrix, 
    \item $\mathbf{b} \in \mathbb{R}^d$ is a learnable bias term,
    \item $\mathbf{h} \in \mathbb{R}^d$ is the resulting embedding representation.
\end{itemize}
Through this linear layer, the input n-dimensional numerical data is projected into the embedding space and subsequently fed into the Transformer encoder.

\subsection{Binning Process}
When continuous numerical data is learned as a regression task, it often fails to achieve satisfactory performance. To address this issue, our method discretizes continuous numerical data, enabling it to be learned as a classification task. This binning process generates the ground-truth labels necessary for self-supervised learning.

The input data $\mathbf{x}_i$ (numerical data for the $i$-th dimension) is divided into an arbitrary number of $k$ bins and converted into discrete labels $y_i$. The binning process is conducted as follows:\\
\noindent
\textbf{Standardizing the numerical scale} : 
For both the Pre-Training dataset and the downstream learning dataset, the numerical data $x_i$ for each dimension $i$ is standardized using the minimum value $x_{\min}$ and maximum value $x_{\max}$:

\[
x_i^{\text{scaled}} = \frac{x_i - x_{\min}}{x_{\max} - x_{\min}}, \quad x_{\min} \leq x_i \leq x_{\max}, \quad 0 \leq x_i^{\text{scaled}} \leq 1
\]
\noindent
\textbf{Discretizing via binning} : 
The standardized data $x_i^{\text{scaled}}$ is divided into $k$ bins, and discrete labels $y_i$ are generated using the floor function:

\[
y_i = \text{min}(\left\lfloor k \cdot x_i^{\text{scaled}} \right\rfloor, k - 1), \quad y_i \in \{0, 1, \dots, k-1\}
\]

This process is performed independently for each dimension $i$ of the input data. By doing so, ground-truth label data for self-supervised learning is generated, which serves as the target data for loss calculation in subsequent sections.

\subsection{Parallel Linear Layers for the Output}
In traditional language Transformer models used for natural language processing, the output layer consists of a single linear layer for predictions. In contrast, our method introduces multiple parallel linear layers, each corresponding to a dimension of the input data. This design allows the representation vectors output by the Transformer encoder to be appropriately processed for each dimension, enabling self-supervised learning.

When the input data consists of $n$-dimensional numerical data, the output layer includes $n$ parallel linear layers. Each linear layer $f_i$ ($i = 1, \dots, n$) corresponds to the $i$-th dimension of the input.

For the representation vector $\mathbf{H} \in \mathbb{R}^d$ obtained from the Transformer encoder, each linear layer $f_i$ is defined as follows:

\[
\hat{y}_i = f_i(\mathbf{H}) = \mathbf{W}_i \mathbf{H} + \mathbf{b}_i \quad (i = 1, 2, \dots, n)
\]

\begin{comment}
    \begin{itemize}
    \item $\mathbf{H} \in \mathbb{R}^d$ : The representation vector from the Transformer encoder,
    \item $\mathbf{W}_i \in \mathbb{R}^{c \times d}$ : The weight matrix of the $i$-th linear layer,
    \item $\mathbf{b}_i \in \mathbb{R}^c$ : The bias term of the $i$-th linear layer,
    \item $\hat{y}_i \in \mathbb{R}^c$ : The output of the $i$-th linear layer,
    \item $c$ : The number of output classes (corresponding to the number of bins in the binning process).
\end{itemize}
\end{comment}

\B{\begin{itemize}
    \item $\mathbf{H} \in \mathbb{R}^d$ is the representation vector from the Transformer encoder.
    \item $\mathbf{W}_i \in \mathbb{R}^{c \times d}$ is the weight matrix of the $i$-th linear layer.
    \item $\mathbf{b}_i \in \mathbb{R}^c$ is the bias term of the $i$-th linear layer.
    \item $\hat{y}_i \in \mathbb{R}^c$ is the output of the $i$-th linear layer.
    \item $c$ is the number of output classes (corresponding to the number of bins in the binning process).
\end{itemize}}

Thus, each linear layer corresponds to a specific dimension of the input data, taking the representation vector $\mathbf{H}$ as input and generating the output for the respective dimension.

During Pre-Training, cross-entropy loss is computed using the labels generated from the binning process. The loss $\mathcal{L}_i$ for each dimension $i$ is expressed as:
\[
\mathcal{L}_i = - \sum_{j=1}^{c} y_{i,j} \log \hat{y}_{i,j}
\]
\begin{comment}
    \begin{itemize}
    \item $y_{i,j}$ : The ground-truth label for class $j$ in the $i$-th dimension (represented as a one-hot vector),
    \item $\hat{y}_{i,j}$ : The predicted probability for the corresponding class (softmax output).
\end{itemize}
\end{comment}
\B{\begin{itemize}
    \item $y_{i,j} \in \{0,1\}$ is the ground-truth label for class $j$ in the $i$-th dimension (represented as a one-hot vector).
    \item $\hat{y}_{i,j} \in [0,1]$ is the predicted probability for the corresponding class (softmax output).
\end{itemize}
}

The final loss function $\mathcal{L}$ is defined as the average loss across all dimensions:
\[
\mathcal{L} = \frac{1}{n} \sum_{i=1}^n \mathcal{L}_i
\]

\subsection{Pre-Training Algorithm}
\begin{algorithm}[H]
    \caption{Pre-training Algorithm}
    \label{algo:pretraining_general}
    \begin{algorithmic}[1]
        \REQUIRE 
            $X \in \mathbb{R}^{(N \times n)}$ \COMMENT{$N$: Number of samples, $n$: n-dimensional sensor data}\\
            \hspace{3em}$\text{model}$; $\text{batch\_size}$; learning rate $\eta$; number of bins $k$; loss function $\mathcal{L}$

        \ENSURE Trained model $\text{model}^*$

        \STATE \COMMENT{\textbf{Step 1: Binning Process for Data}}
        \FOR{$i = 1$ \TO $n$} 
            \STATE $X[:, i]^{\text{scaled}} \gets \frac{X[:, i] - \min(X[:, i])}{\max(X[:, i]) - \min(X[:, i])}$
            \STATE $Y[:, i] \gets \min(\left\lfloor k \cdot X[:, i]^{\text{scaled}} \right\rfloor, k-1)$
        \ENDFOR

        \STATE \COMMENT{\textbf{Step 2: Pretraining the Model}}
        \WHILE{loss $\mathcal{L}$ has not converged}
            \FOR{$X_{\text{batch}}, Y_{\text{batch}}$ in $\text{Loader}(X, Y, \text{batch\_size})$}
                \STATE $Z \gets \text{model}(X_{\text{batch}})$ \COMMENT{Model output}
                \STATE $\mathcal{L} \gets \text{Loss}(Z, Y_{\text{batch}})$
                %\STATE $w \gets w - \eta \cdot \nabla_w \mathcal{L}$
                \STATE $optim.step()$
            \ENDFOR
        \ENDWHILE

        \RETURN $\text{model}^*$
    \end{algorithmic}
\end{algorithm}

The training algorithm using the n-dimensional numerical processing Transformer encoder is presented in Algorithm \ref{algo:pretraining_general}. It is constructed by combining the methods described in Sections 3.1, 3.2, and 3.3. In Step 1, the binning process described in Section 3.2 is applied to generate labels based on the number of bins $k$. In Step 2, the input data is fed into the linear layer described in Section 3.1, and predictions for each dimension's labels are made using the output linear layers described in Section 3.3. The model is trained by updating its parameters to minimize the cross-entropy loss.

This algorithm is not limited to encoder models and can also be applied to decoder models.

\section{Experimental Setup}
In this experiment, we perform Pre-Training and downstream learning using 3-dimensional numerical sensor data. The input data consists of fixed-length sequences of multi-dimensional numerical sensor data (hereafter referred to as "sequences"), which are used to construct both the Pre-Training and the downstream learning task of activity recognition.\\
\noindent
\textbf{What is sequence length?} : 
Sensor data is collected continuously over time, giving each data point a temporal structure. The sequence length refers to the number of data points extracted when segmenting this time-series data into fixed lengths.
For example, when the sequence length is set to 300, one sequence consists of 300 consecutive data points. Each data point is represented as an n-dimensional numerical vector (e.g., the x, y, and z axes of an accelerometer).\\
\noindent
\textbf{Pre-Training} : 
Using a fixed sequence length of 300, we employ the sensor dataset capture24 \cite{Capture24} for Pre-Training.
The proposed method described in Chapter 3, including linear embedding of n-dimensional input data, binning, and parallel linear layers in the output layer, is used to perform Pre-Training with the Transformer encoder (DistilBERT) and Transformer decoder (GPT-2).\\
\noindent
\textbf{Downstream Learning} : 
As the downstream task, activity recognition is performed on five datasets: ADL \cite{adl}, Opportunity \cite{oppotunity}, PAMAP2 \cite{pamap2}, REALWORLD \cite{realworld}, and WISDM \cite{wisdm}.
For each dataset, the data is segmented into sequences of length 300, and the task involves predicting a single activity label for each sequence. The label corresponds to the activity associated with the sensor data at the central point of each sequence.

Through this setup, we evaluate the effectiveness of the sensor data features learned during Pre-Training on the downstream activity recognition tasks. This chapter details the specific experimental settings for Pre-Training.

\subsection{Pre-Training Setup}
\begin{algorithm}[H]
    \caption{Pretraining Algorithm Using Sequence Data}
    \label{algo:pretraining_with_sequence}
    \begin{algorithmic}[1]
        \REQUIRE 
            $X \in \mathbb{R}^{(N \times n)}$ \COMMENT{$N$: Number of samples, $n$: Number of dimensions} \\
            \hspace{3em}Sequence length: 300; Number of bins $k$; Learning rate $\eta$; Model $\text{model}$

        \ENSURE Trained $\text{model}^*$

        \STATE \COMMENT{\textbf{Step 1: Prepare Training Data}}
        \STATE $X \gets \text{normalize}(X)$ \COMMENT{Normalize sensor data}
        \STATE $\text{num\_seq} \gets N // 300$ \COMMENT{Total number of sequences}
        \STATE $X_{\text{seq}} \gets \text{reshape}(X, [\text{num\_seq}, 300, n])$ 
        \STATE $Y_{\text{seq}} \gets \text{discretize}(X_{\text{seq}}, k)$ \COMMENT{Discretize each dimension into bins}

        \STATE \COMMENT{\textbf{Step 2: Pretrain the Model}}
        \WHILE{Loss $\mathcal{L}$ has not converged}
            \FOR{$X_{\text{batch}}, Y_{\text{batch}}$ in $\text{Loader}(X_{\text{seq}}, Y_{\text{seq}})$}
                \STATE $Z \gets \text{model}(X_{\text{batch}})$ \COMMENT{Input sequences into the model}
                \STATE $\mathcal{L} \gets \text{Loss}(Z, Y_{\text{batch}})$ \COMMENT{Calculate loss}
                \STATE $w \gets w - \eta \cdot \nabla_w \mathcal{L}$ \COMMENT{Update parameters using gradients}
            \ENDFOR
        \ENDWHILE

        \RETURN $\text{model}^*$
    \end{algorithmic}
\end{algorithm}

Algorithm \ref{algo:pretraining_with_sequence} outlines the Pre-Training algorithm used in this experiment. Pre-Training is performed using the capture24 dataset, which contains 3-axis numerical data (x, y, z) from accelerometer sensors.

%\noindent
\textbf{Preprocessing} : 
First, the maximum, minimum, and mean values are calculated after removing the top 5\% and bottom 5\% of the data. The top 5\% and bottom 5\% values are then reassigned to the calculated maximum and minimum values, respectively, to mitigate the adverse effects of extreme values on model training.
Next, any missing values are imputed with the mean value. Finally, Min-Max scaling is applied to normalize the data to a range of 0 to 1. This preprocessing is performed independently for each sensor dimension.

%\noindent
\textbf{Input Data Reshaping} : 
The preprocessed sensor data is reshaped into the format (number of data points, sequence length, number of sensor dimensions). This format constructs the data as sequences to be input into the model. The sequence length is set to 300.

%\noindent
\textbf{Data Discretization} : 
The reshaped data is discretized by dividing the values into bins, creating label data for each sensor dimension. For this binning process, two bin sizes (100 and 1000) are used to create the labels. A smaller number of labels simplifies the information to be learned and can enhance generalization performance but risks missing important features. Conversely, a larger number of labels allows the model to learn finer-grained features of the sensor data but may lead to overfitting by over-adapting to sensor measurement noise, potentially reducing generalization performance.

%\noindent
\textbf{Preparation} : 
The prepared data includes fixed-sequence-length sensor data as input and corresponding binned labels. These are used for model input and loss calculation. The output dimensions of the linear layers in the output layer are adjusted to match the number of labels in the data.

%\noindent
\textbf{Training Loop} : 
The prepared data is divided into batches of size 25 and input into the model. Cross-entropy loss is computed between the model output and the label data. The loss is averaged across sequences and dimensions to derive the overall loss, after which model parameters are updated. The AdamW optimization algorithm is used with a learning rate of $5 \times 10^{-5}$.
In this study, the loss converged sufficiently within one epoch, so the training loop was limited to a single epoch.

\subsection{Downstream Learning Setup}
The pretrained model's weights, excluding the classification head, are loaded to perform the activity recognition task.

As in Pre-Training, fixed-length sequences with a sequence length of 300 are used. For each sequence, the activity label corresponding to the central data point is assigned as the activity label for the entire sequence.
The optimization algorithm used is AdamW \cite{AdamW},
with a learning rate of $5 \times 10^{-5}$.

This approach evaluates the performance of the Pre-trained model on the activity recognition task.
Details of the algorithm are provided in the appendix.

\subsection{Setup For Each Pre-Training Task}
In this study, three Pre-Training tasks were used: a reconstruction task, a Masked Language Modeling (MLM) task, and a Next token prediction task. The model architecture used for each Pre-Training method remains consistent; however, modifications were made to the dataset class responsible for creating batches according to the specific requirements of each task.

This section explains the settings of the dataset class used for each Pre-Training task.

\subsubsection{Pre-Training : Reconstruction}
The reconstruction task involves providing multi-dimensional numerical input data to the model and comparing the logits obtained from the model’s output layer with the pre-discretized data to reconstruct the original discretized data. In this experiment, the encoder model DistilBERT was used.

For this task, the loss is computed using cross-entropy between the logits from the output layer and the discretized data. All data points from the logits are included in the loss calculation, and no data points are excluded. The reconstruction task is performed as a classification task across all data points obtained from the model.

\subsubsection{Pre-Training : MLM(Masked Language Modeling)}
The Masked Language Modeling (MLM) task is a Pre-Training task designed to enable the model to learn contextual information by masking a portion of the input sequence and predicting the masked values. In this experiment, the encoder model DistilBERT was used.

For this task, 25\% of the data points are randomly selected and masked across all dimensions and within the sequence range of the input sequence. The masked positions are replaced with a mask value of -100.0. The logits corresponding to unmasked positions are excluded from the cross-entropy calculation, and the loss is computed solely based on the masked positions.

\subsubsection{Pre-Training : Next Token Prediction}
The next-token prediction task is a Pre-Training task primarily performed using decoder models. In this experiment, the decoder model GPT-2 was employed.

The objective of the next-token prediction task is to enable the model to learn the ability to predict the next data point by referencing only the past information up to any given point. In this experiment, the decoder version of the n-dimensional numerical processing Transformer model was used, and the task was designed to predict the multi-dimensional numerical data following a specific data point. The next multi-dimensional numerical data is predicted as discretized classification labels.

For loss calculation, cross-entropy loss was not computed over the entire sequence. Instead, only a specific range was targeted. Specifically, for a sequence length of 300, the first 70 data points in the sequence were excluded from the cross-entropy calculation. Loss calculation for next-token prediction was performed only for the data points following this range.

This configuration was adopted based on the consideration that the early portion of the sequence lacks sufficient contextual information for meaningful next-token prediction. As such, including it in the loss calculation was unlikely to contribute to effective learning.

\section{Result and Discussion}
This chapter presents the experimental results.
\subsection{Assumptions}
For Pre-Training, we utilized the capture24 dataset \cite{Capture24}. For downstream tasks, we employed the ADL \cite{adl}, Opportunity \cite{oppotunity}, PAMAP2 \cite{pamap2}, REALWORLD \cite{realworld}, and WISDM \cite{wisdm} datasets. Additionally, confusion matrices and loss transition plots are presented for further evaluation.

The input data consisted of 3-dimensional accelerometer sensor data (x, y, and z axes) with a sequence length of 300. The task was defined as predicting a single user activity label for each sequence. This task was also used for performance comparison with ResNet \cite{resnet}, Random Forest \cite{randomforest}, and Vanilla Transformer \cite{transformer} models.

Regarding the Vanilla Transformer method shown in Table \ref{table:downstream}, the 3-dimensional sensor data was sequentially arranged as input. This is because the Vanilla Transformer requires a 1-dimensional sequence of tokens as input. For this purpose, the sensor data was first scaled using min-max normalization. Subsequently, all values were scaled up according to a predefined vocabulary size, truncated to discard decimal points, and converted into integer values. These integer values served as token IDs, forming the input token sequence for the Vanilla Transformer. As this experiment focused on 3-dimensional sensor data, the input to the Vanilla Transformer consisted of a 1-dimensional token sequence with a length of 900.

The Pre-Training method employed was a Masked Language Model (MLM) task.
\subsection{Results}
\subsubsection{Intrinsic Evaluation}
\newcommand{\best}[1]{\textbf{\underline{#1}}}

\begin{table}[h]
    \centering
    \caption{Intrinsic Performance Comparison}
    \label{tab:intrinstic_1}
    \adjustbox{width=\textwidth+10em, center}{
    \begin{tabular}{l|cc|cc} 
        \hline
        \multirow{2}{*}{models}&\multicolumn{2}{c}{PAMAP2}&\multicolumn{2}{c}{REALWORLD} \\
        & acc&f1&acc&f1\\
        \hline
        Vanilla Transformer vocab size:10000&0.7126&	0.6940 &0.7479 &0.7446  \\ 
        Vanilla Transformer vocab size:30000&0.6864&	0.6454 	&0.7320 	&0.7252 \\
        \hline\hline

        1-Dimensional Numerical Processing Transformer bin size:100              &0.7758          &0.7721 	   &0.7943              &0.7947 \\
        1-Dimensional Numerical Processing Transformer bin size:1000             &\best{0.7811} 	&\best{0.7777} &	\best{0.7963} 	&\best{0.7969} \\ 
        \hline
    \end{tabular}}
\end{table}

Table \ref{tab:intrinstic_1} presents the results of a comparison between the Vanilla Transformer and our proposed method, using only 1-dimensional acceleration data from the X-axis for training. Both models are based on DistilBERT and were pre-trained on the capture24 dataset. After Pre-Training, downstream tasks were conducted on respective datasets, and performance was evaluated.

The evaluation metrics used were accuracy and weighted F1-score.

In the case of the Vanilla Transformer, performance decreased on both the PAMAP2 and REALWORLD datasets when the vocabulary size was set to 30,000 compared to 10,000. This performance degradation may be attributed to reduced learning efficiency caused by an excessively large vocabulary size.

Conversely, the proposed method outperformed the Vanilla Transformer under both experimental conditions. This indicates that the embedding layer implemented with a linear transformation, the binning process for numerical data, and the output linear layer for classification tasks in the proposed method effectively contributed to feature extraction from the sensor data.

\subsubsection{Extrinsic Evaluation}
%\usepackage{hhline}
% カスタムコマンド
%\newcommand{\best}[1]{\textbf{\underline{#1}}}

\begin{table*}[h]
    \centering
    \caption{Extrinsic Performance Comparison}
    %\hspace{-6cm}    
    \label{table:downstream}
    \adjustbox{width=\textwidth+21em, center}{
\begin{tabular}{l|cc|cc|cc|cc|cc}\hline
\multirow{2}{*}{models} & \multicolumn{2}{c}{ADL} & \multicolumn{2}{c}{Opportunity} & \multicolumn{2}{c}{PAMAP2} & \multicolumn{2}{c}{REALWORLD} & \multicolumn{2}{c}{WISDM} \\
 & acc & f1 & acc & f1 & acc & f1 & acc & f1 & acc & f1 \\ \hline
ResNet18 \cite{resnet}&0.9417&0.9430&0.7960&0.7982&0.8927&0.8912&\best{0.9397}&\best{0.9394}&\best{0.9240}&\best{0.9243}  \\
Random Forest \cite{randomforest}&0.8583&0.8411&0.7143&0.6757&0.7230&0.6910&0.8205&0.8196&0.7329&0.7303 \\
Vanilla Transformer(DistilBERT\cite{distilbert})vocab size:10000 &0.8583&0.8553&0.7272&0.7245&0.7561&0.7527&0.7807&0.7819&0.7923&0.7924 \\
Vanilla Transformer(DistilBERT\cite{distilbert})vocab size:30000 &0.6220&0.6139&0.7002&0.6931&0.6812&0.6595&0.7683&0.7710&0.7123&0.7040 \\
%\R{SignalTransformer-MLM [anonymous]}
SENvT-u4 \cite{Okita23}
&\R{0.9210}&\R{0.8930}&\R{0.7681}&\R{0.7623}&\R{0.8592}&\R{0.8556}&\R{0.9071}&\R{0.9170}&\R{0.8741}&\R{0.8722} \\
SENvT-contrastive \cite{Okita23}
%\R{SignalTransformer-contrastive [anonymous]}
&\R{0.9212}&\R{0.8994}&\R{0.7692}&\R{0.7812}&\R{0.8470}&\R{0.8351}&\R{0.9091}&\R{0.9199}&\R{0.8900}&\R{0.8890} \\
\hline\hline
DistilBERT(Reconstruction)bin size:100 &0.9260&0.9230&0.8166&0.8130&0.8659&0.8659&0.9176&0.9179&\best{0.9130}&\best{0.9127}\\
DistilBERT(Reconstruction)bin size:1000&0.8938&0.8912&0.8021&0.7986&0.8787&0.8791&0.9034&0.9042&0.9082&0.9075  \\ \hline
DistilBERT(MLM)bin size:100&0.9481&0.9469&0.8138&0.8133&0.9025&0.9024&\best{0.9186}&\best{0.9188}&0.9062&0.9058\\
DistilBERT(MLM)bin size:1000&\best{0.9606}&\best{0.9599}&\best{0.8228}&\best{0.8245}&\best{0.9100}&\best{0.9103}&0.9122&0.9127&0.8974&0.8971\\ \hline
GPT2-small(Next Token Prediction)bin size:100&0.9496&0.9491&0.8045&0.8049&0.8840&0.8844&0.9181&0.9184&0.8965&0.8967\\
GPT2-small(Next Token Prediction)bin size:1000&0.9307&0.9275&0.8122&0.8079&0.8745&0.8751&0.9088&0.9091&0.8964&0.8963\\ \hline
GPT2-medium(Next Token Prediction)bin size:100&0.9417&0.9379&0.8021&0.8027&0.8686&0.8679&0.9177&0.9178&0.8959&0.8961\\
GPT2-medium(Next Token Prediction)bin size:1000&0.9528&0.9510&0.8115&0.8079&0.8885&0.8888&0.9112&0.9113&0.8865&0.8871\\
\hline
DistilBERT (without Pre-Trainig)&0.8567& 	0.8484& 	0.7693& 	0.7587& 	0.8188& 	0.8187& 	0.8957& 	0.8964& 	0.8856& 	0.8853\\
GPT2-small (without Pre-Trainig)&0.8708& 	0.8553& 	0.7175& 	0.6970& 	0.8111& 	0.8104& 	0.8525& 	0.8534& 	0.8435& 	0.8437\\
GPT2-midium (without Pre-Trainig)&0.8504& 	0.8159& 	0.7032& 	0.6588& 	0.7805& 	0.7778& 	0.8454& 	0.8466& 	0.8630& 	0.8636 
\end{tabular}
}
    \caption*{Note: ResNet18, Random Forest, and Vanilla Transformer are employed as baseline models. In contrast, models based on DistilBERT and GPT-2 incorporate the proposed methods. The table compares their performance under various pre-training methods as well as cases without pre-training.}

\end{table*}
For comparison, baseline models including ResNet, Random Forest, and Vanilla Transformer were employed. Additionally, the proposed methods, based on DistilBERT and GPT-2, were evaluated. The results include the performance of various pre-training methods as well as cases without pre-training.

When comparing the performance of the Vanilla Transformer and the proposed methods, the proposed methods outperformed the Vanilla Transformer across all datasets. In the Vanilla Transformer, continuous numerical data is converted into discrete token IDs, which are then transformed into representation vectors through an embedding layer. This process can potentially disrupt the continuity and relationships inherent in the numerical data.

Furthermore, due to the architectural design of the Vanilla Transformer, the embedding layer only accepts discrete token IDs as input, making it incapable of directly processing multidimensional sensor data as a single data point. This limitation complicates handling multidimensional data obtained from multiple sensor axes. In contrast, the proposed method uses a linear embedding layer, enabling the model to input multidimensional data as a single data point. This fundamental difference significantly impacts the ability to process sensor data effectively. These architectural limitations and the difficulty of representing small continuous changes in multidimensional sensor data using discrete token IDs likely contributed to the lower performance of the Vanilla Transformer.

On the ADL, Opportunity, and PAMAP2 datasets, the proposed methods outperformed ResNet and Random Forest. Particularly on the Opportunity dataset, the proposed methods achieved higher accuracy and F1 scores than ResNet for all pre-training methods. However, on the REALWORLD and WISDM datasets, ResNet outperformed the proposed methods, and none of the pre-training methods surpassed ResNet's performance.

Within the proposed methods, reconstruction tasks achieved the highest performance only on the WISDM dataset. On the other datasets, the Masked Language Modeling (MLM) task consistently recorded the best performance, suggesting that the MLM task is relatively effective for training language transformer-based models in the proposed framework.

Pre-training using the next-token prediction task did not show any significant advantage over the other two methods in downstream tasks.

For the discretization of sensor data during pre-training, comparisons were made using bin sizes of 100 and 1000. However, no clear impact of bin size on performance was observed in this experiment.

\R{
%SignalTransformer-MLM [anonymous]\footnote{Self reference.} and SignalTransformer-contrastive [anonymous]\footnote{Self reference.} 
SENvT-u4 and SENvT-contrastive \cite{Okita23}
outperform the Vanilla Transformer in five downstream tasks. This demonstrates that the mechanism of treating one-dimensional patches as embeddings in the SENvT
%SignalTransformer 
is effective. However, DistilBERT (MLM) outperformed both SENvT-u4 and SENvT-contrastive
%SignalTransformer-MLM and SignalTransformer-contrastive 
in all five downstream tasks. This indicates that our DistilBERT (MLM) performed even better, suggesting that the proposed approach leads to high performance. As is mentioned in Section 2.3, 
%SignalTransformer-MLM 
SENvT-u4 handles patches while our DistilBert (MLM) handles each data point projected in a linear manner.}
%% DOUBLE BLIND \cite{Okita23}

\subsubsection{Analysis of Oppotunity}
\begin{figure}[H]
    \centering
    \includegraphics[width=0.9\textwidth]{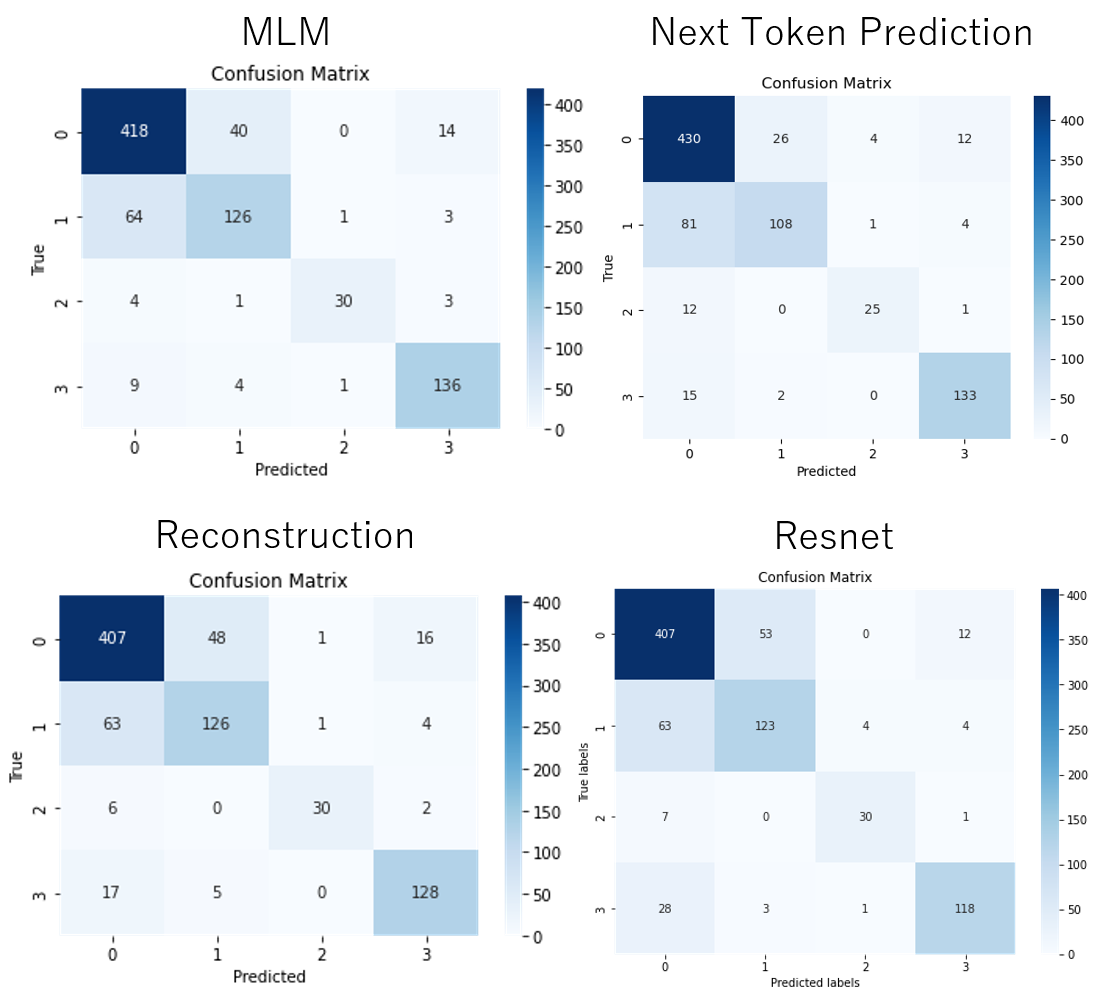}
    \caption{Confusion matrix for Oppotunity}
    \label{fig:confusion_matrix_oppo}
\end{figure}
On the Opportunity dataset, downstream models pre-trained using the Masked Language Modeling (MLM) task achieved the highest accuracy, with improvements of up to 2.68\% in accuracy and 2.63\% in F1-score compared to other methods. These results surpassed those of both the reconstruction task and the next-token prediction task. Moreover, the performance of all proposed methods exceeded that of ResNet.

The confusion matrix in Figure \ref{fig:confusion_matrix_oppo} illustrates the performance of downstream models trained with different pre-training tasks, as well as the performance of the pre-trained ResNet model on the test data.

While the proposed method using the MLM task demonstrated superior performance on the Opportunity dataset, the reasons for this high performance require further investigation. Specifically, comparative analyses with various other tasks represent an important direction for future research.

\subsubsection{Analysis of RealWorld}
\begin{figure}[H]
    \centering
    \includegraphics[width=0.9\textwidth]{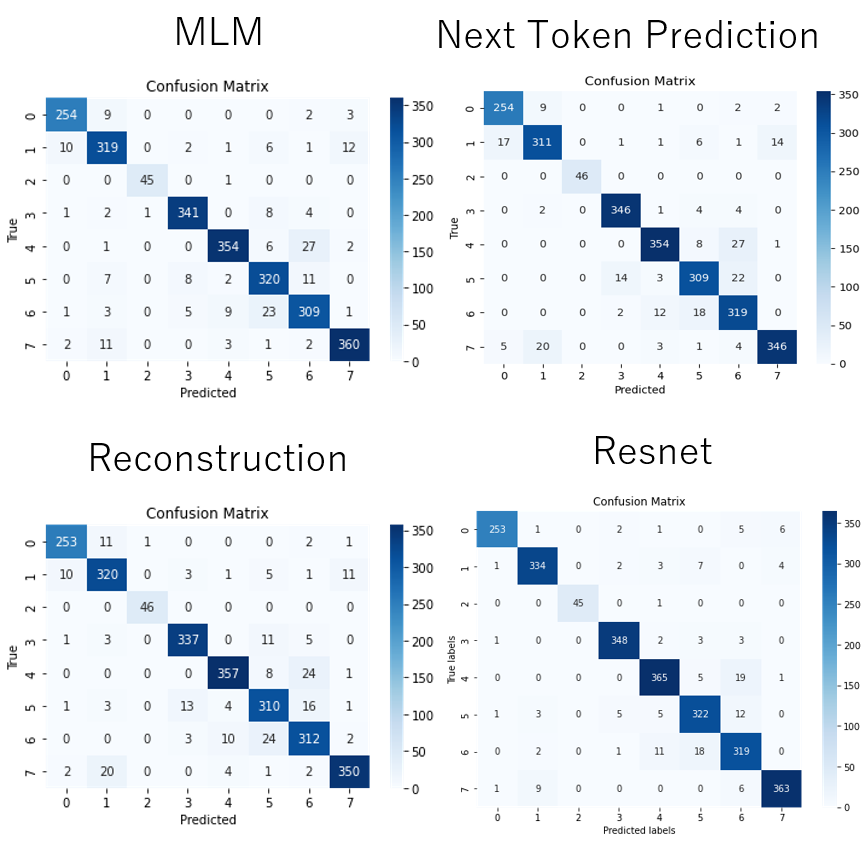}
    \caption{Confusion matrix for Oppotunity}
    \label{fig:confusion_matrix_realworld}
\end{figure}
On the REALWORLD dataset, none of the proposed pre-training tasks outperformed ResNet. However, among the proposed methods, pre-training with the Masked Language Modeling (MLM) task achieved the highest performance, surpassing the other tasks.

Both the Opportunity and REALWORLD datasets demonstrated the effectiveness of the MLM task compared to other tasks. However, while all proposed methods outperformed ResNet on the Opportunity dataset, none of them surpassed ResNet on the REALWORLD dataset. This discrepancy likely stems from differences in the characteristics of the two datasets.

A detailed examination of the differences between the Opportunity and REALWORLD datasets is necessary to determine whether the proposed pre-training tasks themselves require improvement, whether the architecture needs refinement, or under what conditions the proposed methods are most effective. This represents an important direction for future research.

\subsubsection{Pre-Training Loss Values}
\begin{figure}[H]
    \centering
    \begin{adjustbox}{width=\textwidth+10em, center}
        \includegraphics{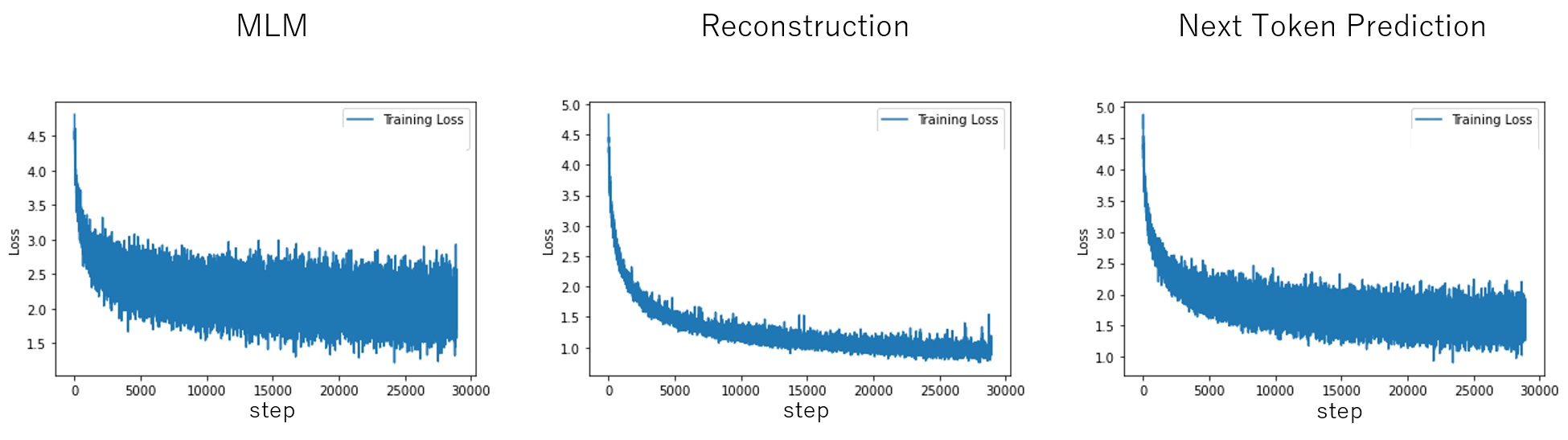}
    \end{adjustbox}
    \caption{Confusion matrix for Oppotunity}
    \label{fig:pretrain_loss}
\end{figure}
Figure \ref{fig:pretrain_loss} illustrates the loss transitions for each pre-training task, specifically for a bin size of 100. The initial random loss value was approximately 4.61. The final training and validation losses for each task are summarized in Table \ref{tab:pretrain_loss}.
\begin{table}[h]
    \centering
    \caption{Final Loss Values for Pre-Training Tasks}
    \label{tab:pretrain_loss}
    \begin{tabular}{lcc} 
        \hline
        \textbf{Pre-Training task} & \textbf{Train Loss} & \textbf{Validation Loss} \\ 
        \hline
        MLM              & 2.2396              & 2.3242             \\ 
        Reconstruction           & \textbf{1.2649}     & \textbf{1.0327}    \\ 
        Next-token Prediction    & 1.8121              & 1.8213             \\ 
        \hline
    \end{tabular}
\end{table}
The MLM task exhibited the most unstable loss reduction, with significant fluctuations during training. Although its final loss values were higher compared to other tasks, it still demonstrated a certain level of convergence. The reconstruction task showed the most stable loss reduction and achieved the lowest final loss values, both for training and validation, indicating high generalization performance on evaluation data. The next-token prediction task yielded intermediate results, with final loss values that did not match those of the reconstruction task.

From the earlier performance evaluation table, downstream task performance followed the trend: MLM task \R{$>$} Reconstruction task \R{$>$} Next-token prediction task. Despite the reconstruction task achieving the most stable and lowest pre-training losses, its performance in downstream tasks did not surpass the MLM task. This suggests that minimizing pre-training task losses alone does not directly translate to improved downstream performance. It is hypothesized that the reconstruction task might have overly adapted to its specific pre-training task, hindering its ability to effectively enhance downstream task performance.

In contrast, the MLM task, with its more unstable convergence, may have avoided overfitting to the pre-training task. This likely allowed it to learn more effective features of the sensor data for downstream tasks. Therefore, it can be concluded that setting a pre-training task that is neither overly simplistic nor excessively complex is crucial when employing n-dimensional numerical data with Transformer models.

\section{\R{Model Comparison}}
%\usepackage{hhline}
% カスタムコマンド
%\newcommand{\best}[1]{\textbf{\underline{#1}}}

\begin{table}[h]
    \centering
    \caption{model comparison of parameters}
    %\hspace{-6cm}    
    \label{table:comparison_params}
    \adjustbox{width=\textwidth+15em, center}{
\begin{tabular}{cccccc}\hline
Model & Model Parameters(Pre-training) &  Model Parameters(Downstream) & hidden\_size & n\_heads& n\_layers\\
\hline
ResNet18 &-&3,848,904&-&-&-  \\
DistilBERT bin size:100&44,768,556& 	44,355,905& 	768 &	12 &	6  \\
GPT2-small bin size:100&85,522,476& 	85,297,928& 	768 &	12 &	12\\

\hline

\end{tabular}
}
\caption*{Model Parameters : This value represents the total number of trainable parameters during both the pre-training and downstream learning phases, indicating the total number of parameters updated throughout the model}
\end{table}
%\usepackage{hhline}
% カスタムコマンド
%\newcommand{\best}[1]{\textbf{\underline{#1}}}

\begin{table}[h]
    \centering
    \caption{model comparison of training/inferece times\\ and memory consumption}
    %\hspace{-6cm}    
    \label{table:comparison_time}
    \adjustbox{width=\textwidth+13em, center}{
\begin{tabular}{ccccccc}\hline
Model &  
\begin{tabular}{c}
     Pre-training\\ (min)
\end{tabular}
&   
\begin{tabular}{c}
     Downstream\\ (sec)
\end{tabular}
& 
\begin{tabular}{c}
     Inference on \\GPU(ms/sample)
\end{tabular}
& 
\begin{tabular}{c}
     Inference on \\CPU(ms/sample)
\end{tabular}
& 
\begin{tabular}{c}
     Memory \\Usage (MB)
\end{tabular}
\\
\hline
ResNet18 &-&5	&2.2950 	&4.5370 &	578.4800  \\
DistilBERT bin size:100& 86 	&71 	&4.5420 	&74.0620 	&957.6400  \\
GPT2-small bin size:100& 175 	&300 	&6.3373 	&155.0240 	&1401.9800  \\

\hline

\end{tabular}
}
\caption*{Note: Memory Usage is the amount of memory used when measured by inference in the CPU. Also training and inference times are presented using the Capture24 dataset for pretraining and the PAMAP2 dataset for downstream learning.}
\end{table}
\R{
The following discussion is based on the results presented in Tables \ref{table:comparison_params} and \ref{table:comparison_time}, considering the requirements of a practical real-time HAR system. In this setting, the sensor device operates at 30 fps, meaning that each sample, consisting of 300 sensor sequences, must be processed within approximately 33 ms. First, ResNet18 was trained only on the downstream task without any pretraining, requiring only 5 seconds for training. In contrast, the proposed Transformer-based models underwent both pretraining and downstream task learning. Specifically, DistilBERT required 86 minutes and GPT-2 required 175 minutes for pretraining, while the downstream training times were 71 seconds and 300 seconds, respectively.
}
Regarding inference time, all models demonstrated significantly faster inference than the 30 fps requirement ($\leq$33 ms per sample) in the GPU setting, with ResNet18 achieving 2.30 ms per sample, DistilBERT 4.54 ms per sample, and GPT-2\_small 6.34 ms per sample. However, in the CPU setting, ResNet18 remained highly efficient at 4.54 ms per sample, whereas DistilBERT and GPT-2\_small exhibited considerably longer inference times of 74.06 ms per sample and 155.02 ms per sample, respectively. These results indicate that Transformer-based models are impractical for direct deployment in edge environments where GPU acceleration is unavailable.
%%%
In terms of memory footprint
%%%
(Table \ref{table:comparison_time}:RAM usage),
%%%
ResNet18 required approximately 578 MB, whereas DistilBERT and GPT-2\_small consumed 958 MB and 1402 MB, respectively.
This substantial increase in memory usage for Transformer-based models is attributed to their multi-layer self-attention mechanisms and large intermediate representations.
However, the proposed method demonstrated notable performance improvements through pretraining.
In our experiments, it outperformed ResNet18 on three out of five datasets, indicating that the computational and memory costs are justified by the accuracy gains.
Thus, while all models satisfy the 30 fps real-time processing requirement under GPU conditions, ResNet18 remains the most advantageous choice for CPU-based real-time HAR applications.
%%%%%%
Overall, Transformer-based models exhibit challenges in terms of training cost, CPU inference speed, and memory consumption. However, their ability to achieve superior accuracy through pretraining represents a significant advantage. On the other hand, ResNet18 requires no pretraining, has a lower training cost, and outperforms Transformer-based models in inference speed and memory efficiency, making it the most practical choice for real-time HAR deployment on edge devices.

\section{Conclusion}
\label{section:Conclusion}
\begin{comment}
    In this study, we proposed a pre-training method using an n-dimensional numerical Transformer model and applied it to activity recognition tasks based on sensor data. The proposed method incorporated three types of pre-training tasks: Masked Language Modeling (MLM), Reconstruction, and Next-token Prediction, and their performance was compared.

The proposed method outperformed conventional approaches on the ADL, Opportunity, and PAMAP2 datasets, with the MLM task achieving the highest performance. Additionally, no direct correlation was observed between the reduction in pre-training task losses and downstream performance, as the MLM task exhibited the most unstable loss reduction but achieved the best overall performance.

Future work involves refining pre-training methods by adjusting the difficulty of the MLM task for performance comparison and investigating the relationship between dataset characteristics and pre-training tasks. This will help in developing more effective pre-training techniques.
\end{comment}
In this study, we proposed a pre-training method utilizing an n-dimensional numerical Transformer model and applied it to activity recognition tasks based on sensor data. The proposed method incorporated three types of pre-training tasks—Masked Language Modeling (MLM), Reconstruction, and Next-Token Prediction—whose performances were systematically compared.

Our method outperformed conventional approaches on the ADL, Opportunity, and PAMAP2 datasets, with the MLM task achieving the highest performance. 
\R{Additionally, we compared our method with similar Transformer-based models, including SENvT-u4 and SENvT-contrastive.
%SignalTransformer-MLM and SignalTransformer-contrastive. 
While these models outperformed the Vanilla Transformer across multiple downstream tasks, our DistilBERT (MLM) achieved even better results, demonstrating the effectiveness of handling each data point through a linear projection rather than patch-based embeddings.}
Notably, no direct correlation was observed between the reduction in pre-training task losses and downstream performance. Despite exhibiting the most unstable loss reduction, the MLM task yielded the best overall results.

For future work, we aim to refine the pre-training approach by adjusting the difficulty of the MLM task for performance comparison and further investigating the relationship between dataset characteristics and pre-training task effectiveness. These insights will contribute to the development of more robust and efficient pre-training techniques.

\bibliographystyle{plain}
\bibliography{bibtex}

\appendix
% ここでデータセットのtrain, validation, testの配分表みたいなものを表示するといいかも。あとダウンストリーム学習アルゴリズムも
\section{Appendix}
\subsection{Dataset}
%\usepackage{hhline}
% カスタムコマンド
%\newcommand{\best}[1]{\textbf{\underline{#1}}}

\begin{table*}[h]
    \centering
    \caption{Train, Validation, and Test Sample Counts for Datasets}
    %\hspace{-6cm}    
    \label{table:dataset}
    \adjustbox{width=\textwidth, center}{
\begin{tabular}{ccccc}\hline
Dataset & Classes &  Train Samples & Validation Samples & Test Samples \\
\hline
ADL\cite{adl}&	5	&406	&102&	127 \\
Oppotunity\cite{oppotunity}&	4	&2730	&683	&854\\
PAMAP2\cite{pamap2}	&8	&1836	&459	&574\\
REALWORLD\cite{realworld}&	8	&7964	&1992	&2490\\
WISDM\cite{wisdm}	&18	&17916	&4480	&5600\\
Capture-24\cite{Capture24} &- & 722308 & 187227 & -\\
\hline

\end{tabular}
}
\end{table*}
Table \ref{table:dataset} summarizes the five datasets used for the extrinsic evaluation presented in Table \ref{table:downstream}. The column "Class" represents the number of activity types in the activity recognition task. The columns "Train Samples," "Validation Samples," and "Test Samples" show the number of samples allocated for training, validation, and testing, respectively, in each dataset. These datasets consist of 3-dimensional time-series data, where one sample corresponds to a sequence of 300 timesteps. This format is used as input for each model in the study.

Additionally, the capture-24 dataset, used for pretraining, is included in the table. Since this dataset is not used for the activity recognition task, the "Class" column is not applicable in this case. Only the "Train Samples" and "Validation Samples" are shown for this dataset.

\subsection{Downstream Algorithm}
\begin{algorithm}[H]
    \caption{Downstream Learning}
    \label{algo:downstream}
    \begin{algorithmic}[1]
        \REQUIRE 
            $X_{\text{seq}} \in \mathbb{R}^{(\text{num\_seq} \times \text{sequence\_length} \times \text{sensor\_dim})}$\\
            \hspace{3.2em}$Y_{\text{label}} \in \mathbb{R}^{(\text{num\_seq})}$\\
            \hspace{3.2em}Pretrained $\text{model}$; batch size $\text{batch\_size}$; \\
            \hspace{3.2em}Learning rate $\eta$

        \ENSURE Trained $\text{model}^*$

        \WHILE{the training has not converged}
            \FOR{$X_{\text{batch}}, Y_{\text{batch}}$ in $\text{Loader}(X_{\text{seq}}, Y_{\text{label}}, \text{batch\_size})$}
                \STATE $Z \gets \text{model}(X_{\text{batch}})$
                \STATE $\mathcal{L} \gets \text{CrossEntropy}(Z, Y_{\text{batch}})$
                \STATE $w \gets w - \eta \cdot \nabla_w \mathcal{L}$ \COMMENT{Update parameters based on the loss}
            \ENDFOR
        \ENDWHILE
        
        \RETURN $\text{model}^*$
    \end{algorithmic}
\end{algorithm}

This algorithm \ref{algo:downstream} describes the downstream learning process conducted after self-supervised learning in the proposed method. The input data is structured as three-dimensional sensor data, consistent with the pretraining phase, where each sample corresponds to a sequence of 300 timesteps. Using this data, activity recognition is performed as a classification task.

During downstream learning, the pretrained model is loaded for the activity recognition task. In this process, the weights of the parallel linear layers in the output layer are discarded. A new classification head for activity recognition is introduced, while all other parameters are loaded from the pretrained model. This approach leverages the knowledge acquired during pretraining to construct a model tailored for the activity recognition task.

\end{document}